\documentclass{article}

\usepackage{microtype}
\usepackage{graphicx}
\usepackage{subcaption}
\usepackage{booktabs} 
\usepackage{enumitem}

\usepackage{hyperref}

\usepackage[accepted]{icml2026_weightsymmetry}

\usepackage{amsmath}
\usepackage{amssymb}
\usepackage{mathtools}
\usepackage{amsthm}
\usepackage{caption}

\usepackage[capitalize,noabbrev]{cleveref}

\theoremstyle{plain}
\newtheorem{theorem}{Theorem}[section]
\newtheorem{proposition}[theorem]{Proposition}

\theoremstyle{definition}

\theoremstyle{remark}

\usepackage[textsize=tiny]{todonotes}

\icmltitlerunning{Different Layers, Different Manifolds}

\begin{document}

\twocolumn[
  \icmltitle{Different Layers, Different Manifolds: \\
  Module-Wise Weight-Space Geometry in Transformer Optimization}



  \icmlsetsymbol{equal}{*}

  \begin{icmlauthorlist}
    \icmlauthor{Kirato Yoshihara}{uosaka}
  \end{icmlauthorlist}

  \icmlaffiliation{uosaka}{School of Engineering Science, The University of Osaka, Osaka, Japan}
  \icmlcorrespondingauthor{Kirato Yoshihara}{kiratoyoshihara@gmail.com}

  \icmlkeywords{weight-space geometry, manifold optimization, Manifold Muon, training dynamics}

  \vskip 0.3in
]



\printAffiliationsAndNotice{}  

\begin{abstract}
    Weight-space geometry plays a central role in neural network optimization, yet manifold constraints are often applied uniformly across all weight matrices. In this work, we ask whether different transformer modules prefer different manifold geometries. We study Manifold Muon for GPT-2 pretraining and compare layer-wise assignments of Stiefel and DGram constraints across attention and MLP blocks. Our results show a clear asymmetry: constraining attention layers with Stiefel geometry while assigning DGram geometry to MLP layers gives the best performance among the tested configurations, whereas the inverted assignment and all-DGram configuration become unstable under the shared hyperparameter setting. We trace this failure to singular value growth in DGram-constrained attention weights, which can amplify attention logits and induce softmax saturation. These findings suggest that symmetry-aware and geometry-aware optimization for transformers should be module-specific rather than uniform. Code is available at \url{https://github.com/kiratoyoshihara/module-wise-manifold-muon}.
\end{abstract}

\section{Introduction}

Neural networks exhibit rich weight-space geometries and symmetries, arising from permutation, scaling, and rotation-like degrees of freedom in their parameterizations \citep{entezari2022the, ainsworth2023git}. These structures affect optimization trajectories, training dynamics, and the spectral behavior of learned weights. Recent optimizers such as Muon exploit this structure through matrix-normalized updates to hidden-layer weights \citep{jordan2024muon}, with subsequent work demonstrating scalability to large language model training and emphasizing the role of update scale and weight decay choices \citep{liu2025muon}. Recent analysis also connects Muon-style updates to spectral-norm constraints and singular-value behavior in Transformer projections \citep{chen2025muon}.

Manifold Muon and related analyses further frame this direction as optimization under structured matrix-manifold constraints, including Stiefel-type geometries \citep{bernstein2025manifolds, cesista2025spectralclipping}. More recent work has also explored manifold optimization directly for LLM training \citep{gu2026mano}. These works suggest that weight-space geometry is not merely a theoretical artifact, but a practical design axis for improving stability and shaping training dynamics in modern networks. Unlike these works, we focus not on proposing a new manifold optimizer, but on how different manifold geometries should be assigned across transformer modules.

Prior work on orthogonality and manifold-constrained training commonly studies a single constraint family, such as orthogonality or Stiefel constraints, applied across collections of neural network weight matrices \citep{orthogonal,feedback,yang2026manifold}. This leaves open a basic question for transformers \citep{attention}: do different module types prefer the same weight-space geometry? Attention and MLP layers play different computational roles, and their weights enter the network through different nonlinear mechanisms. In particular, weaker Gram-space constraints such as DGram preserve some structure while allowing additional scale degrees of freedom \citep{keigwin2025gram}. Such freedom may be beneficial in some modules but harmful in others.

In this work, we study this question in GPT-2 pretraining \citep{gpt2} by comparing layer-wise assignments of Stiefel and DGram constraints across attention and MLP blocks. Our main finding is a clear asymmetry under a shared training configuration: assigning Stiefel geometry to attention layers and DGram geometry to MLP layers gives the best performance among the tested configurations, whereas the inverted assignment and the all-DGram configuration become unstable. Thus, uniform manifold constraints are insufficient; transformer optimization benefits from module-specific weight-space geometry. We do not claim that DGram attention can never be stabilized; rather, our results show that, without additional scale control or retuned hyperparameters, placing DGram on attention exposes an unstable singular-value growth pathway.

We further analyze why DGram attention becomes unstable in this setting. Because DGram permits singular value growth, attention weight products can amplify logits, pushing the softmax into a saturated regime and potentially degrading training dynamics. This provides a mechanistic explanation for the observed training instability and suggests a practical principle: attention layers require spectrally bounded geometry, while MLP layers can benefit from weaker scale-preserving freedom.

Our contributions are threefold. First, we evaluate module-wise manifold assignments for Manifold Muon in GPT-2 pretraining. Second, we show that attention and MLP layers prefer different weight-space geometries. Third, we identify singular value growth and softmax saturation as a plausible instability mechanism for DGram-constrained attention.

\section{Background and Setup}
\subsection{Manifold Muon}

Muon applies matrix-normalized updates to hidden-layer weight matrices, modifying the geometry of the update rather than treating each parameter independently \citep{jordan2024muon}. Manifold Muon extends this view by constraining weight matrices to structured matrix manifolds during optimization \citep{bernstein2025manifolds}. 

We focus on two constraint families. The first is the Stiefel constraint, which enforces column orthonormality,
\[
    W^\top W = I .
\]
This bounds the singular values of the constrained matrix and therefore imposes strong spectral control. The second is DGram, a weaker Gram-space constraint that requires
the Gram matrix to be diagonal but does not fix its diagonal
entries:
\[
    \operatorname{Off}(W^\top W) = 0,
    \quad \text{i.e.}, \quad
    (W^\top W)_{ij} = 0 \text{ for } i \neq j .
\]
Unlike Stiefel, DGram does not fix the column norms and therefore
permits larger singular values, providing additional scale degrees
of freedom \citep{keigwin2025gram}.

\subsection{Layer-wise Manifold Assignments}

Transformers contain heterogeneous modules with distinct computational roles, most notably attention and MLP/FFN blocks \citep{attention}. We therefore ask whether these module types should share the same manifold geometry. We partition the constrained weight matrices into attention weights and MLP/FFN weights, and assign either Stiefel or DGram constraints to each group.

We evaluate five assignments: \textsc{Unconstrained}, with no manifold constraint; \textsc{All-Stiefel}, with Stiefel constraints for both attention and MLP/FFN weights; \textsc{All-DGram}, with DGram constraints for both groups; \textsc{Hetero}, with Stiefel attention and DGram MLP/FFN weights; and \textsc{Hetero-Inv}, with DGram attention and Stiefel MLP/FFN weights.

This setup directly tests whether a single manifold geometry suffices for transformer training, or whether attention and MLP layers prefer different weight-space geometries.

\begin{center}
\small
\captionof{table}{Validation loss and training outcome for each assignment. ``Unst.'' denotes an unstable run under the operational criterion in Appendix~\ref{app:details}.}
\label{tab:main_results}
\begin{tabular}{lllc}
\toprule
Assignment & Attn. & MLP & Result \\
\midrule
\textsc{Uncon.} & None & None & 3.3855 \\
\textsc{All-St.} & Stiefel & Stiefel & 3.3679 \\
\textsc{All-DG.} & DGram & DGram & Unst. \\
\textsc{Hetero} & Stiefel & DGram & \textbf{3.3544} \\
\textsc{Het.-Inv} & DGram & Stiefel & Unst. \\
\bottomrule
\end{tabular}
\end{center}

\section{Different Layers Prefer Different Manifolds}

\subsection{Experimental Setup}

We evaluate the layer-wise manifold assignments in GPT-2 small pretraining using a nanoGPT-style implementation \citep{gpt2, karpathy2022nanogpt}. The model has approximately 124M parameters and is trained on OpenWebText \citep{Gokaslan2019OpenWeb}.

All configurations use the same model architecture, data, training schedule, and optimizer hyperparameters except for the manifold assignment. This isolates the effect of assigning different weight-space geometries to attention and MLP/FFN modules. Full implementation details and hyperparameters are provided in the appendix.

\subsection{Main Result}

\begin{figure*}[t]
    \centering
    \includegraphics[width=0.85\linewidth]{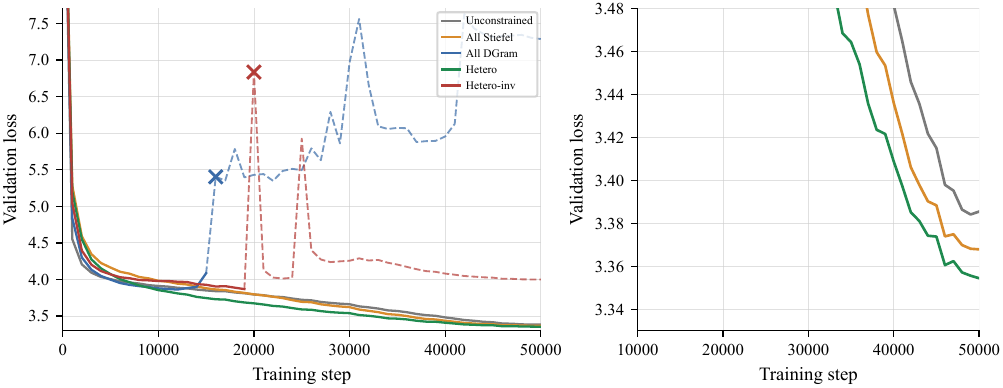}
    \caption{
    Validation loss curves for the five layer-wise manifold assignments.
    Left: full training trajectory, including unstable runs. Right: zoomed view
    of the stable configurations in the late-training regime.
    \textsc{Hetero} achieves the lowest stable validation loss, while both
    configurations that assign DGram to attention, \textsc{All-DGram} and
    \textsc{Hetero-Inv}, become unstable.
    }
    \label{fig:val_loss_curves}
\end{figure*}

Table~\ref{tab:main_results} summarizes the final validation loss and training outcome for each assignment. The best stable configuration is \textsc{Hetero}, which assigns Stiefel constraints to attention weights and DGram constraints to MLP/FFN weights. It improves over both the unconstrained baseline and the uniform \textsc{All-Stiefel} assignment.

The more important observation is qualitative: in our experiments, every configuration that places DGram on attention becomes unstable under the shared hyperparameter setting. Both \textsc{All-DGram} and \textsc{Hetero-Inv} become unstable, whereas configurations with Stiefel attention remain stable. This suggests that, under this optimizer setting, attention layers require stronger spectral control than MLP/FFN layers. Conversely, DGram appears compatible with MLP/FFN weights and can even improve performance when paired with Stiefel-constrained attention.

Figure~\ref{fig:val_loss_curves} shows the corresponding validation loss curves. The stable runs follow similar early training trajectories, but \textsc{Hetero} reaches the lowest final validation loss. In contrast, the DGram-attention configurations exhibit unstable dynamics and fail to recover.

\section{Mechanistic Analysis: Why DGram Attention Fails}

\subsection{Singular Value Growth in Attention}

We next analyze why the assignments that place DGram on attention become unstable in our shared hyperparameter setting. The key empirical difference is spectral: DGram attention permits rapid growth of the largest singular value, whereas Stiefel attention keeps the constrained matrices spectrally bounded by construction. Since attention logits depend on products of query and key projections, uncontrolled growth in these matrices can directly amplify the scale of the attention logits.

\begin{figure*}[t]
    \centering
    \includegraphics[width=0.90\linewidth]{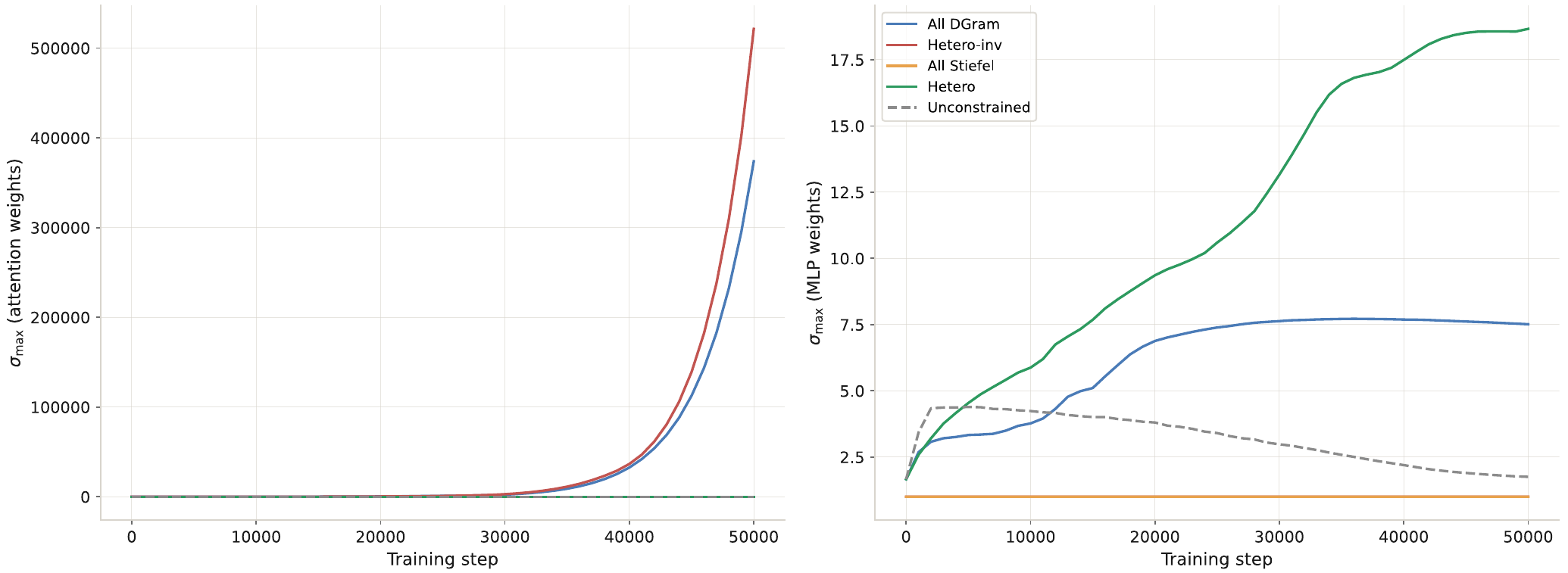}
    \caption{
    Spectral evolution during training. Each curve reports the maximum
    singular value over all matrices in the corresponding parameter
    group across layers. Left: attention weights. DGram attention exhibits explosive singular value growth in \textsc{All-DGram} and \textsc{Hetero-Inv} under the shared hyperparameter setting. Right: MLP/FFN weights. DGram is compatible
    with MLP/FFN weights and grows without becoming unstable when
    attention remains Stiefel-constrained.
    }
    \label{fig:spectral_evolution}
\end{figure*}

Figure~\ref{fig:spectral_evolution} shows the evolution of the largest singular value during training. In the left panel, configurations with DGram attention, \textsc{All-DGram} and \textsc{Hetero-Inv}, exhibit explosive growth in $\sigma_{\max}$ for attention weights, reaching extremely large values before the runs become unstable. In contrast, configurations with Stiefel attention keep the attention spectrum bounded. This matches the stability pattern in Figure~\ref{fig:val_loss_curves}: the runs with DGram attention are precisely those classified as unstable.

The right panel shows a contrasting pattern for MLP/FFN
weights: under HETERO, DGram MLP weights grow in spectral
norm yet training remains stable. We return to this
asymmetry in Section~\ref{sec:mlp-tolerates}.

\subsection{Softmax Saturation and Gradient Degeneracy}

To explain this module-specific failure mode, consider a single attention head with input token representations $X \in \mathbb{R}^{T \times d_{\mathrm{model}}}$, where $T$ is the sequence length. Let $W_Q, W_K \in \mathbb{R}^{d_{\mathrm{model}} \times d}$ denote the query and key projection matrices for a head of dimension $d$. The pre-softmax attention logits are
\[
    Z
    =
    \frac{X W_Q W_K^\top X^\top}{\sqrt{d}}
    \in \mathbb{R}^{T \times T}.
\]
Thus, the scale of $Z$ is directly affected by the spectral norms of $W_Q$ and $W_K$. In particular, for any fixed input $X$,
\[
    \|Z\|_2
    \leq
    \frac{\|X\|_2^2 \, \|W_Q\|_2 \, \|W_K\|_2}{\sqrt{d}} .
\]
This inequality does not prove instability, but it shows why singular value growth in attention projections can amplify the attention logits.

Large logit magnitudes push the softmax into a saturated regime.
Let $z \in \mathbb{R}^T$ denote one row of $Z$ and
$p = \operatorname{softmax}(z)$, with Jacobian
$J_{\mathrm{softmax}}(z) = \operatorname{diag}(p) - pp^\top$.
If the largest logit exceeds the rest by a gap $\Delta > 0$, then
$1 - p_1 \leq (T-1)e^{-\Delta}$, so $p$ approaches a one-hot vector
exponentially fast and $\|J_{\mathrm{softmax}}(z)\|_F$ becomes small.

\paragraph{Gradient non-degeneracy.}
The preceding bounds can be stated as a structural contrast between
Stiefel and DGram attention.

\begin{proposition}[Stiefel attention is gradient non-degenerate under bounded inputs]
\label{prop:stiefel-nondeg}
Let $W_Q,W_K \in \mathrm{St}(d_{\mathrm{model}},d)$ and assume bounded input
features $\|x_i\|_2 \leq R$ for all tokens $i$. Then there exists a constant
$c(T,d,R)>0$, independent of $W_Q$ and $W_K$, such that for every row $z$
of the attention logit matrix $Z$,
\[
    \|J_{\mathrm{softmax}}(z)\|_F \geq c(T,d,R).
\]
\end{proposition}

\begin{proposition}[DGram attention admits gradient-degenerate directions under bounded inputs]
\label{prop:dgram-deg}
Under the DGram constraint, there exist feasible query and key projection
matrices whose largest singular values grow arbitrarily large. Consequently,
for some bounded input features and for every $\varepsilon>0$, the induced
attention logits can produce a row $z$ such that
\[
    \|J_{\mathrm{softmax}}(z)\|_F < \varepsilon .
\]
\end{proposition}

Proof sketches are given in Appendix~\ref{app:proofs}.
Together, these formalize the asymmetry in
Figure~\ref{fig:spectral_evolution}, providing a plausible
mechanism for unstable training: singular value growth
$\to$ logit inflation $\to$ softmax saturation. This is a mechanistic explanation, not a complete proof that DGram attention is inherently unstable under all optimizer settings.

\subsection{Why MLPs Tolerate DGram}
\label{sec:mlp-tolerates}

The same scale freedom appears less harmful in MLP/FFN layers because the nonlinearity has a different structure from attention softmax. In a transformer MLP block, the hidden activations are applied coordinate-wise, e.g.,
\[
    \operatorname{MLP}(x)
    =
    W_{\mathrm{out}} \, \phi(W_{\mathrm{in}}x),
\]
where $\phi$ is typically GELU. Unlike softmax attention, this operation does not couple all token-to-token scores through a normalized probability simplex. Saturation or amplification in one coordinate therefore does not force the entire module into a nearly one-hot routing regime.

Moreover, GELU acts element-wise and its derivative remains locally defined by each coordinate rather than by global competition among logits. In addition, transformer blocks use LayerNorm and residual connections, which can partially absorb changes in activation scale before the next block. These architectural differences help explain why DGram-induced singular value growth can be tolerated in MLP/FFN weights, while the same freedom is destabilizing in attention projections.

\section{Conclusion and Limitations}

Our results suggest that manifold constraints for transformer optimization should be assigned module-wise rather than uniformly. Attention layers appear to require the spectral control provided by Stiefel constraints, whereas MLP/FFN layers can tolerate, and in our experiments benefit from, the weaker scale freedom of DGram constraints. This supports the broader view that weight-space geometry is a practical design axis for transformer training.

There are several limitations. Our experiments are limited to GPT-2 small on OpenWebText and a small number of training runs. In addition, all manifold assignments are compared under a shared hyperparameter setting. This controls for the training configuration, but it does not rule out the possibility that DGram attention could be stabilized by retuning the learning rate, adding explicit scale control, or applying weight decay to DGram-managed weights. Therefore, the modest validation-loss gap between \textsc{Hetero} and \textsc{All-Stiefel} should be interpreted cautiously and validated with additional seeds and larger scales. However, the qualitative instability of DGram-attention configurations is consistent across both \textsc{All-DGram} and \textsc{Hetero-Inv}, and is supported by the observed singular value growth in attention weights.

Our mechanistic analysis is also not a complete proof of training instability. It shows that singular value growth in attention projections can amplify attention logits and push softmax into a saturated regime, but future work should directly measure logit scales, attention entropy, and gradient flow throughout training. More broadly, an important direction is to develop automatic methods for assigning manifold constraints across modules, architectures, and scales.




\section*{Impact Statement}
This work studies optimizer geometry for transformer
pretraining and has no foreseeable societal impact beyond
that of foundational ML research.


\bibliography{references}
\bibliographystyle{icml2026}

\newpage
\appendix
\onecolumn

\section{Experimental Details}
\label{app:details}

\paragraph{Model and data.}
We train a GPT-2 small--style Transformer language model from scratch on OpenWebText using a nanoGPT-style implementation. Table~\ref{tab:model_data_details} summarizes the model and data configuration.

\begin{table}[h]
\centering
\caption{Model and data configuration.}
\label{tab:model_data_details}
\begin{tabular}{ll}
\toprule
Item & Value \\
\midrule
Model & GPT-2 small--style Transformer \\
Training initialization & From scratch \\
Dataset & OpenWebText \\
Data loader & nanoGPT-style memmap loader \\
Objective & Next-token prediction \\
Number of layers & 12 \\
Number of attention heads & 12 \\
Embedding dimension & 768 \\
Context length & 1024 \\
Dropout & 0.0 \\
Linear layer bias & Disabled \\
Vocabulary size & 50{,}304 (GPT-2 BPE, padded for efficiency) \\
\bottomrule
\end{tabular}
\end{table}

\paragraph{Training setup.}
All configurations use the same training schedule and batching setup, summarized in Table~\ref{tab:training_details}.

\begin{table}[h]
\centering
\caption{Training configuration.}
\label{tab:training_details}
\begin{tabular}{ll}
\toprule
Item & Value \\
\midrule
Training iterations & 50{,}000 \\
Batch size & 64 \\
Gradient accumulation steps & 1 \\
Tokens per optimization step & 65{,}536 \\
Evaluation interval & Every 1{,}000 iterations \\
Evaluation batches & 200 \\
Random seed & 1337 \\
Mixed precision & \texttt{bfloat16} \\
Compilation & \texttt{torch.compile} \\
Gradient clipping & Global norm 1.0 \\
\bottomrule
\end{tabular}
\end{table}

\paragraph{Optimizer setup.}
We use a hybrid optimizer setup: attention and MLP projection matrices are updated with Muon or Manifold Muon, while the remaining parameters are updated with AdamW \citep{loshchilov2019decoupled}. Table~\ref{tab:optimizer_assignment} shows the parameter assignment, and Table~\ref{tab:optimizer_hparams} summarizes the optimizer hyperparameters.

\begin{table}[h]
\centering
\caption{Parameter assignment to optimizers.}
\label{tab:optimizer_assignment}
\begin{tabular}{lll}
\toprule
Parameter group & Name pattern & Optimizer \\
\midrule
Attention projections & \texttt{attn.c\_attn}, \texttt{attn.c\_proj} & Muon / Manifold Muon \\
MLP projections & \texttt{mlp.c\_fc}, \texttt{mlp.c\_proj} & Muon / Manifold Muon \\
Other 2D parameters & Embeddings, LM head, etc. & AdamW with weight decay \\
1D parameters & LayerNorm, scales, etc. & AdamW without weight decay \\
\bottomrule
\end{tabular}
\end{table}

\begin{table}[h]
\centering
\caption{Optimizer hyperparameters.}
\label{tab:optimizer_hparams}
\begin{tabular}{lll}
\toprule
Optimizer & Hyperparameter & Value \\
\midrule
AdamW & Peak learning rate & $6 \times 10^{-4}$ \\
AdamW & Betas & $(0.9, 0.95)$ \\
AdamW & Weight decay (2D params)   & $0.1$ \\
AdamW & Weight decay (1D params)   & $0$   \\
Muon (unconstrained) & Weight decay & $0.1$ \\
Manifold Muon & Weight decay & Not applied to manifold-managed weights \\
Muon / Manifold Muon & Peak learning rate & $0.02$ \\
Muon / Manifold Muon & Momentum & $0.95$ \\
Manifold Muon & Dual-ascent steps & 20 \\
Manifold Muon & Dual-ascent step size & $0.1$ \\
\bottomrule
\end{tabular}
\end{table}

For Stiefel-managed weights, the manifold itself fixes singular values and provides implicit scale control. For DGram-managed weights, the diagonal of $W^\top W$ is unconstrained, so no explicit scale control is applied in our experiments. This is a potential confound: weight decay or other scale regularization may stabilize DGram attention.

\paragraph{Learning-rate schedule.}
Both AdamW and Muon use the same cosine learning-rate schedule with linear warmup. The warmup period is 1{,}000 iterations, the cosine decay runs until 50{,}000 iterations, and the minimum AdamW learning rate is $6 \times 10^{-5}$. The Muon learning rate is scaled proportionally to the AdamW schedule: if $\eta_t^{\mathrm{AdamW}}$ is the AdamW learning rate at iteration $t$, then the Muon learning rate is
\[
\eta_t^{\mathrm{Muon}}
=
0.02 \cdot
\frac{\eta_t^{\mathrm{AdamW}}}{6 \times 10^{-4}}.
\]
Thus, AdamW and Muon share the same warmup and decay profile while retaining different peak learning rates.




\paragraph{Manifold Muon implementation.}
We implement Manifold Muon as a tangent-space-constrained Muon update
followed by a retraction. For each Muon-managed matrix
$W \in \mathbb{R}^{m \times n}$, we internally transpose wide matrices
so that $m \ge n$ before applying the operations below, and transpose
back at the end. Let $G$ denote the Nesterov momentum direction.

\paragraph{Tangent constraint.}
We characterize each manifold by a self-adjoint projector
$P_{\mathcal{M}}$ acting on $n \times n$ symmetric matrices. The
tangent condition takes the unified form
\[
P_{\mathcal{M}}(W^\top A + A^\top W) = 0,
\]
where $P_{\mathcal{M}}$ selects the components of the symmetric Gram
differential that the manifold actually constrains. For Stiefel, the
constraint $W^\top W = I$ controls every entry of the Gram matrix, so
$P_{\mathrm{St}}$ is the identity on symmetric matrices: the entire
symmetric part must vanish. For DGram, we constrain only the off-diagonal entries of the Gram
matrix. The projector is
\[
P_{\mathrm{DG}}(M) = M - \mathrm{Diag}(\mathrm{diag}(M)),
\]
which selects the off-diagonal part of a symmetric matrix. The
resulting tangent condition
\[
\mathrm{Off}(W^\top A + A^\top W) = 0
\]
preserves $\mathrm{Off}(W^\top W) = 0$, i.e., the columns of $W$
remain pairwise orthogonal. The diagonal entries of $W^\top W$,
which correspond to per-column scales (and hence to singular values
since the columns are orthogonal), are unconstrained. DGram is
therefore weaker than Stiefel: it preserves Gram diagonality but
does not force $W^\top W = I$.

The update direction $A$ is computed by approximately solving
\[
\max_{A \in \mathbb{R}^{m \times n}}
\mathrm{tr}(G^\top A)
\quad
\mathrm{s.t.}
\quad
\|A\|_2 \le 1,
\quad
P_{\mathcal{M}}(W^\top A + A^\top W) = 0,
\]
where $\|\cdot\|_2$ denotes the spectral norm. The corresponding
parameter step is then $W \mapsto W - \eta A$, with $\eta$ the Muon
learning rate.

\paragraph{Dual ascent.}
We solve this problem by dual ascent over a multiplier
$\Lambda \in \mathbb{R}^{n \times n}$. Since the constraint depends
only on the symmetric part of $\Lambda$, we pass
$\tfrac{1}{2}(\Lambda + \Lambda^\top)$ through $P_{\mathcal{M}}$ in
the gradient computation; this also keeps the dual update
well-defined when $\Lambda$ drifts away from symmetry during
iteration. The Lagrangian is linear in $A$ apart from the
spectral-norm bound, and its maximizer over the unit spectral-norm
ball is given by the matrix sign function:
\[
\arg\max_{\|A\|_2 \le 1} \mathrm{tr}(B^\top A) = \mathrm{msign}(B).
\]
Here $\mathrm{msign}(B) = UV^\top$ for the compact SVD
$B = U\Sigma V^\top$, equivalently the polar factor of $B$. Applying this to the gradient of the Lagrangian, the inner-loop
update direction at multiplier $\Lambda_k$ is
\[
A_k =
\mathrm{msign}
\!\left(
G + 2 W \, P_{\mathcal{M}}
\!\left(
\tfrac{\Lambda_k + \Lambda_k^\top}{2}
\right)
\right),
\]
where the factor $2$ arises from the derivative of
$W^\top A + A^\top W$ with respect to $A$.

We initialize the multiplier with
\[
\Lambda_0 = -\tfrac{1}{4}\, P_{\mathcal{M}}(W^\top G + G^\top W),
\]
which we use as a simple linearized warm start for the dual ascent iterations. In practice, this reduces the number of inner iterations needed to obtain a small tangent-constraint violation.

At inner iteration $k$, we measure the constraint violation
\[
H_k = P_{\mathcal{M}}(W^\top A_k + A_k^\top W)
\]
and update the multiplier as
\[
\Lambda_{k+1} = \Lambda_k
- \alpha\!\left(1 - \tfrac{k}{K}\right) H_k.
\]
The factor $(1 - k/K)$ anneals the dual step size linearly to zero so
that the iterates settle within $K$ inner steps rather than
oscillating around the constraint surface. We use $K = 20$ and
$\alpha = 0.1$ in all experiments.

\paragraph{Retraction.}
After the inner loop, we form the trial update
\[
\widetilde{W} = W - \eta A_K
\]
and retract it back onto the assigned manifold. For Stiefel, the
retraction is the polar factor, computed via the matrix sign:
\[
R_{\mathrm{St}}(\widetilde{W}) = \mathrm{msign}(\widetilde{W}).
\]
For DGram, let the compact SVD of $\widetilde{W}$ be
\[
\widetilde{W} = U \Sigma V^\top,
\qquad
\Sigma = \mathrm{Diag}(\sigma_1, \ldots, \sigma_n).
\]
We use the SVD-based retraction
\[
R_{\mathrm{DG}}(\widetilde{W}) = (U V^\top)\, \Sigma.
\]
Since $(UV^\top)^\top (UV^\top) = I$, the retracted matrix satisfies
\[
R_{\mathrm{DG}}(\widetilde{W})^\top R_{\mathrm{DG}}(\widetilde{W})
=
\Sigma^2,
\]
which is diagonal but not in general the identity. The retraction
therefore eliminates off-diagonal Gram correlations while leaving
the diagonal entries (the squared singular values) free, giving
DGram its characteristic scale freedom relative to Stiefel.

This ensures that each constrained weight starts on its assigned
manifold; subsequent updates are followed by the corresponding
retraction.

\paragraph{Evaluation and logging.}
We evaluate training and validation loss every 1{,}000 iterations. Each evaluation loss is computed as the mean over 200 mini-batches sampled from the corresponding split. We also log per-iteration training loss, wall-clock time per step, learning rate, and model-flop utilization (MFU) estimates for monitoring. Evaluation metrics are saved as JSONL files for later plotting.

\paragraph{Spectral diagnostics.}
For Muon-managed parameters in the manifold-constrained
configurations, we additionally save spectral statistics every
$1{,}000$ iterations. For each such weight matrix, we compute and
store its singular values, the Gram diagonal
$\operatorname{diag}(W^\top W)$, the Frobenius norm of the
off-diagonal part of $W^\top W$, and the Frobenius norm $\|W\|_F$.

For Figure~\ref{fig:spectral_evolution}, we aggregate the spectral
logs by taking the maximum singular value over all matrices in each
parameter group at each logging step. The attention curve aggregates
over \texttt{attn.c\_attn} and \texttt{attn.c\_proj} across all
layers, while the MLP/FFN curve aggregates over \texttt{mlp.c\_fc}
and \texttt{mlp.c\_proj}. We report the maximum rather than the mean
because the softmax-saturation mechanism depends on worst-case
spectral growth.

\paragraph{Instability criterion.}
We classify a run as unstable if any of the following objective conditions are met during the 50,000-iteration schedule:
\begin{enumerate}[leftmargin=2em,topsep=0.2em,itemsep=0.1em]
\item \textit{Hard loss spike.} After iteration $1{,}000$, the
      validation loss exceeds $\max(2 \cdot \mathrm{loss}_{\mathrm{best}},
      5.0)$ at any evaluation step.
\item \textit{Persistent drift.} The mean validation loss over the
      final $20\%$ of evaluations exceeds
      $1.2 \cdot \mathrm{loss}_{\mathrm{best}}$, indicating that the
      run has departed from its best loss and not recovered.
\item \textit{Failure to recover from degradation.} After reaching
      $\mathrm{loss}_{\mathrm{best}}$, the validation loss stays
      above $1.05 \cdot \mathrm{loss}_{\mathrm{best}}$ for the
      remainder of training.
\item \textit{Early termination.} The run halts due to non-finite
      loss before iteration $48{,}000$.
\end{enumerate}
Configurations that meet any of these criteria are reported as unstable in Table~\ref{tab:main_results}. Small fluctuations within
the stable trajectory of converged runs are not treated as
unstable.

\section{Manifold Assignment Schematic}

The five manifold configurations are defined by which constraint
(Stiefel, DGram, or none) is applied to the attention and MLP/FFN
parameter groups; the parameter grouping is given in
Table~\ref{tab:optimizer_assignment} of Appendix~A, and the resulting
configurations and their final validation losses are summarized in
Table~\ref{tab:main_results} of the main body. All non-Muon
parameters use AdamW with the same hyperparameters in every
configuration.

The HETERO configuration is the main heterogeneous assignment, with
Stiefel applied to attention and DGram applied to MLP/FFN. HETERO-INV
swaps these assignments and serves as a control: it tests whether
the benefit comes from the specific Stiefel-attention/DGram-MLP
pairing rather than from heterogeneity per se.

\begin{figure}[t]
\centering
\includegraphics[width=0.5\linewidth]{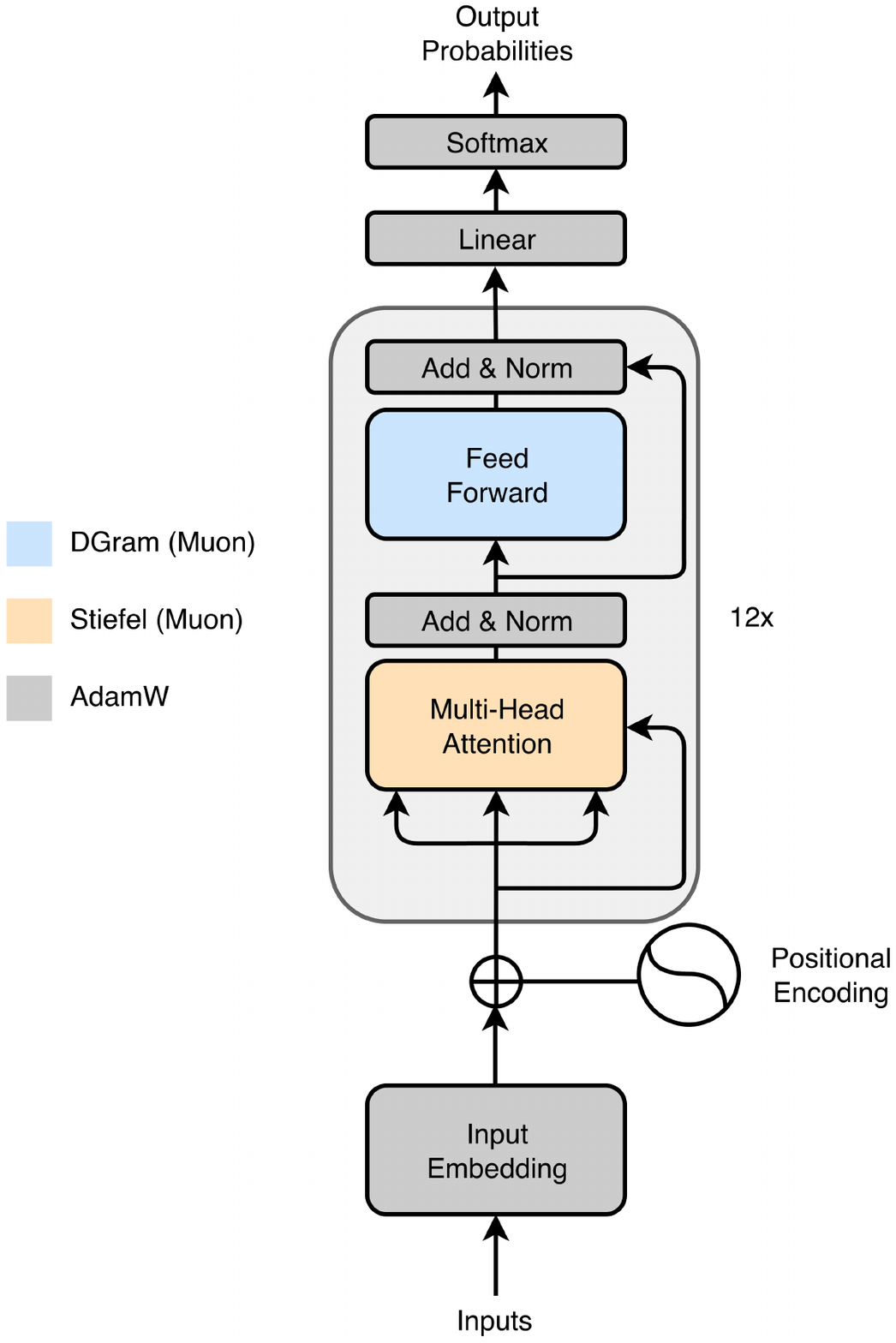}
\caption{
Schematic of the HETERO manifold assignment in GPT-2. Within each
of the 12 Transformer blocks, attention projections are updated by
Stiefel Manifold Muon (orange) and MLP/FFN projections by DGram
Manifold Muon (blue). All remaining parameters---input embedding,
positional encoding, final linear layer, and normalization
scales---are optimized with AdamW (gray).
}
\label{fig:hetero_assignment}
\end{figure}

Figure~\ref{fig:hetero_assignment} illustrates the HETERO assignment
in the full GPT-2 architecture. Each of the 12 Transformer blocks
applies Stiefel Manifold Muon to its attention projections and DGram
Manifold Muon to its feed-forward projections, while all parameters
outside these blocks (input embedding, positional encoding, final
linear layer, layer-norm scales) are optimized with AdamW.

\section{Proof Sketches}
\label{app:proofs}

\paragraph{Spectral consequences of Stiefel and DGram constraints.}
Let \(W \in \mathbb{R}^{m \times n}\) with \(m \ge n\), and let
\(\sigma_1(W) \ge \cdots \ge \sigma_n(W) \ge 0\) denote its singular
values. Under the Stiefel constraint,
\[
W^\top W = I_n .
\]
Therefore, for every \(v \in \mathbb{R}^n\),
\[
\|Wv\|_2^2
=
v^\top W^\top W v
=
\|v\|_2^2 .
\]
Hence all singular values are equal to one:
\[
\sigma_i(W)=1
\qquad
\text{for all } i=1,\ldots,n,
\]
and in particular
\[
\|W\|_2 = \sigma_1(W)=1 .
\]
Thus Stiefel-constrained weights are spectrally bounded by construction.

DGram imposes a weaker Gram constraint. In our implementation, DGram
constrains the off-diagonal entries of the Gram matrix,
\[
\operatorname{Off}(W^\top W)=0,
\] 
so that
\[
W^\top W = D
\]
for some diagonal matrix \(D \succeq 0\). Writing
\[
D=\operatorname{Diag}(d_1,\ldots,d_n),
\]
we have, for every \(v \in \mathbb{R}^n\),
\[
\|Wv\|_2^2
=
v^\top D v
=
\sum_{j=1}^n d_j v_j^2 .
\]
Therefore the singular values of \(W\) are
\[
\sigma_j(W)=\sqrt{d_j},
\qquad j=1,\ldots,n.
\]
Unlike Stiefel, DGram does not require \(d_j=1\). Consequently,
\[
\|W\|_2
=
\max_j \sqrt{d_j}
\]
is unconstrained by the manifold definition. In particular, for any
\(c>0\), the matrix
\[
W_c = c
\begin{bmatrix}
I_n \\
0
\end{bmatrix}
\in \mathbb{R}^{m \times n}
\]
satisfies
\[
\operatorname{Off}(W_c^\top W_c)=0,
\qquad
W_c^\top W_c = c^2 I_n,
\]
but
\[
\|W_c\|_2=c .
\]
Thus DGram preserves Gram diagonality while permitting arbitrary
singular-value growth.

\paragraph{Attention logit amplification.}
Consider a single attention head with token representation matrix
\(X \in \mathbb{R}^{T \times d_{\mathrm{model}}}\), query projection
\(W_Q \in \mathbb{R}^{d_{\mathrm{model}} \times d}\), and key
projection \(W_K \in \mathbb{R}^{d_{\mathrm{model}} \times d}\).
The pre-softmax attention logit matrix is
\[
Z
=
\frac{X W_Q W_K^\top X^\top}{\sqrt d}
\in \mathbb{R}^{T \times T}.
\]
By submultiplicativity of the spectral norm,
\[
\|Z\|_2
=
\left\|
\frac{X W_Q W_K^\top X^\top}{\sqrt d}
\right\|_2
\le
\frac{
\|X\|_2
\|W_Q\|_2
\|W_K^\top\|_2
\|X^\top\|_2
}{\sqrt d}.
\]
Using
\[
\|W_K^\top\|_2=\|W_K\|_2,
\qquad
\|X^\top\|_2=\|X\|_2,
\]
we obtain
\[
\|Z\|_2
\le
\frac{
\|X\|_2^2
\|W_Q\|_2
\|W_K\|_2
}{\sqrt d}.
\]
Thus the scale of the attention logits can grow multiplicatively with
the spectral norms of the query and key projections.

Under Stiefel constraints on both \(W_Q\) and \(W_K\), we have
\[
\|W_Q\|_2=\|W_K\|_2=1,
\]
and therefore
\[
\|Z\|_2
\le
\frac{\|X\|_2^2}{\sqrt d}.
\]
Thus Stiefel attention rules out weight-induced amplification of the
attention logits.

Under DGram constraints, however, write
\[
W_Q^\top W_Q = D_Q,
\qquad
W_K^\top W_K = D_K,
\]
where \(D_Q,D_K\) are diagonal but not necessarily identity. Then
\[
\|W_Q\|_2 = \sqrt{\lambda_{\max}(D_Q)},
\qquad
\|W_K\|_2 = \sqrt{\lambda_{\max}(D_K)}.
\]
Consequently,
\[
\|Z\|_2
\le
\frac{
\|X\|_2^2
\sqrt{\lambda_{\max}(D_Q)}
\sqrt{\lambda_{\max}(D_K)}
}{\sqrt d}.
\]
Since \(\lambda_{\max}(D_Q)\) and \(\lambda_{\max}(D_K)\) are not fixed
by the DGram constraint, DGram attention permits weight-induced logit
amplification.

The same conclusion can be seen at the level of individual logits. For
tokens \(x_i,x_j \in \mathbb{R}^{d_{\mathrm{model}}}\),
\[
Z_{ij}
=
\frac{x_i^\top W_Q W_K^\top x_j}{\sqrt d}.
\]
By Cauchy--Schwarz,
\[
|Z_{ij}|
\le
\frac{
\|x_i\|_2
\|W_Q\|_2
\|W_K\|_2
\|x_j\|_2
}{\sqrt d}.
\]
Thus, even when token features are bounded, large singular values in
\(W_Q\) or \(W_K\) can enlarge the range of the attention logits.

\paragraph{Softmax saturation.}
Let \(z \in \mathbb{R}^T\) be a row of the attention logit matrix and
let
\[
p = \operatorname{softmax}(z),
\qquad
p_i = \frac{e^{z_i}}{\sum_{\ell=1}^T e^{z_\ell}} .
\]
Assume without loss of generality that \(z_1=\max_i z_i\), and define
the logit gap
\[
\Delta
=
z_1 - \max_{j\ne 1} z_j .
\]
Then
\[
1-p_1
=
\frac{\sum_{j\ne 1} e^{z_j}}{e^{z_1}+\sum_{j\ne 1}e^{z_j}}
\le
\frac{\sum_{j\ne 1} e^{z_j}}{e^{z_1}}.
\]
Since \(z_j \le z_1-\Delta\) for all \(j\ne 1\),
\[
\sum_{j\ne 1} e^{z_j}
\le
(T-1)e^{z_1-\Delta}.
\]
Therefore
\[
1-p_1
\le
(T-1)e^{-\Delta}.
\]
Equivalently,
\[
p_1 \ge 1-(T-1)e^{-\Delta}.
\]
Thus, as the logit gap \(\Delta\) grows, the softmax output approaches a
one-hot vector exponentially fast.

The Jacobian of the softmax is
\[
J_{\mathrm{sm}}(z)
=
\frac{\partial p}{\partial z}
=
\operatorname{Diag}(p)-pp^\top .
\]
For any \(u \in \mathbb{R}^T\),
\[
u^\top J_{\mathrm{sm}}(z) u
=
\sum_{i=1}^T p_i u_i^2
-
\left(\sum_{i=1}^T p_i u_i\right)^2
=
\operatorname{Var}_{i \sim p}(u_i)
\ge 0,
\]
so \(J_{\mathrm{sm}}(z)\) is positive semidefinite. Moreover,
\[
J_{\mathrm{sm}}(z)_{ii}
=
p_i(1-p_i),
\qquad
J_{\mathrm{sm}}(z)_{ij}
=
-p_i p_j
\quad (i\ne j).
\]
If \(p_1\) is close to one, write
\[
\varepsilon = 1-p_1.
\]
Then
\[
\sum_{j\ne 1} p_j = \varepsilon.
\]
The Frobenius norm satisfies
\[
\|J_{\mathrm{sm}}(z)\|_F^2
=
\sum_{i=1}^T p_i^2(1-p_i)^2
+
\sum_{i\ne j} p_i^2p_j^2 .
\]
The first diagonal term is
\[
p_1^2(1-p_1)^2
\le
\varepsilon^2.
\]
For \(j\ne 1\),
\[
p_j^2(1-p_j)^2 \le p_j^2,
\]
and hence
\[
\sum_{j\ne 1} p_j^2(1-p_j)^2
\le
\sum_{j\ne 1} p_j^2
\le
\left(\sum_{j\ne 1}p_j\right)^2
=
\varepsilon^2.
\]
For the off-diagonal terms,
\[
\sum_{i\ne j} p_i^2p_j^2
=
\left(\sum_i p_i^2\right)^2 - \sum_i p_i^4
\le
2p_1^2\sum_{j\ne 1}p_j^2
+
\left(\sum_{j\ne 1}p_j^2\right)^2
\le
2\varepsilon^2+\varepsilon^4.
\]
Combining these bounds gives
\[
\|J_{\mathrm{sm}}(z)\|_F^2
\le
4\varepsilon^2+\varepsilon^4,
\]
and therefore
\[
\|J_{\mathrm{sm}}(z)\|_F
\le
2\varepsilon+\varepsilon^2.
\]
Using \(\varepsilon = 1-p_1 \le (T-1)e^{-\Delta}\), we obtain
\[
\|J_{\mathrm{sm}}(z)\|_F
\le
2(T-1)e^{-\Delta}
+
(T-1)^2e^{-2\Delta}.
\]
Thus large logit gaps make the softmax Jacobian exponentially small.
This formalizes the saturation mechanism: if spectral growth in
attention projections increases logit gaps, the softmax can enter a
nearly one-hot regime in which gradients through the attention
probabilities become small.

\paragraph{Proof of Proposition~\ref{prop:dgram-deg}.}
We now combine the preceding observations to show that DGram attention
can realize arbitrarily saturated softmax maps even when the input
features are bounded. This is not a statement about all DGram weights,
but rather an existence result showing that the DGram constraint alone
does not prevent gradient-degenerate attention.

Let \(m=d_{\mathrm{model}}\), and assume \(m \ge d\). Consider DGram-feasible
query and key matrices of the form
\[
W_Q = c
\begin{bmatrix}
I_d \\
0
\end{bmatrix},
\qquad
W_K = c
\begin{bmatrix}
I_d \\
0
\end{bmatrix},
\]
for some scale \(c>0\). Then
\[
W_Q^\top W_Q = c^2 I_d,
\qquad
W_K^\top W_K = c^2 I_d,
\]
so
\[
\operatorname{Off}(W_Q^\top W_Q)=0,
\qquad
\operatorname{Off}(W_K^\top W_K)=0.
\]
Thus \(W_Q\) and \(W_K\) are feasible under the DGram constraint, while
\[
\|W_Q\|_2=\|W_K\|_2=c .
\]

Now choose bounded token features \(x_1,\ldots,x_T\) such that their first
\(d\) coordinates contain distinguishable components. For example, take
\[
x_1 =
\begin{bmatrix}
e_1 \\
0
\end{bmatrix},
\qquad
x_j =
\begin{bmatrix}
0 \\
0
\end{bmatrix}
\quad
\text{for } j\ne 1,
\]
where \(e_1\in\mathbb{R}^d\). These inputs satisfy
\[
\|x_i\|_2 \le 1
\qquad
\text{for all } i.
\]
For the first query row, the logits are
\[
z_j
=
\frac{x_1^\top W_Q W_K^\top x_j}{\sqrt d}.
\]
Hence
\[
z_1
=
\frac{c^2}{\sqrt d},
\qquad
z_j=0
\quad
\text{for } j\ne 1.
\]
The logit gap is therefore
\[
\Delta = \frac{c^2}{\sqrt d}.
\]
By the softmax saturation bound above,
\[
\|J_{\mathrm{sm}}(z)\|_F
\le
2(T-1)\exp\!\left(-\frac{c^2}{\sqrt d}\right)
+
(T-1)^2\exp\!\left(-\frac{2c^2}{\sqrt d}\right).
\]
Consequently, for every \(\varepsilon>0\), there exists a sufficiently
large \(c\) such that
\[
\|J_{\mathrm{sm}}(z)\|_F < \varepsilon .
\]
Thus DGram-feasible attention projections can induce arbitrarily small
softmax Jacobians on bounded inputs. In this precise sense, DGram
attention can become gradient-degenerate.

More generally, the construction does not rely on the specific basis
choice above. If DGram allows
\[
W_Q^\top W_Q = D_Q,
\qquad
W_K^\top W_K = D_K,
\]
with large diagonal entries, then query-key inner products can be scaled
by the corresponding singular directions. Since DGram does not fix the
diagonal of the Gram matrix, it does not rule out this amplification
mechanism.

\paragraph{Proof of Proposition~\ref{prop:stiefel-nondeg}.}
In contrast, Stiefel constraints remove the scale degree of freedom in
the query and key projections. If
\[
W_Q^\top W_Q = I_d,
\qquad
W_K^\top W_K = I_d,
\]
then
\[
\|W_Q\|_2=\|W_K\|_2=1.
\]
For any bounded token features satisfying
\[
\|x_i\|_2 \le R
\qquad
\text{for all } i,
\]
each individual attention logit satisfies
\[
|Z_{ij}|
=
\left|
\frac{x_i^\top W_Q W_K^\top x_j}{\sqrt d}
\right|
\le
\frac{
\|x_i\|_2
\|W_Q\|_2
\|W_K\|_2
\|x_j\|_2
}{\sqrt d}
\le
\frac{R^2}{\sqrt d}.
\]
Therefore all logits lie in the interval
\[
Z_{ij}\in
\left[
-\frac{R^2}{\sqrt d},
\frac{R^2}{\sqrt d}
\right].
\]
For any row \(z\) of \(Z\), its maximum possible logit gap is bounded by
\[
\Delta(z)
=
\max_i z_i - \max_{j\ne i^\star} z_j
\le
\max_i z_i - \min_j z_j
\le
\frac{2R^2}{\sqrt d}.
\]
Thus Stiefel attention prevents the weights themselves from producing
arbitrarily large attention-logit gaps when the input features are
bounded.

Equivalently, at the matrix level,
\[
\|Z\|_2
\le
\frac{\|X\|_2^2}{\sqrt d}.
\]
This does not imply that softmax saturation is impossible, since
\(\|X\|_2\) and the data-dependent geometry of the token features can
still affect the logits. It does imply, however, that Stiefel removes
the specific failure mode in which the query and key projection matrices
grow in spectral norm and amplify the logits through
\[
\|W_Q\|_2\|W_K\|_2 .
\]
This is the sense in which Stiefel attention controls weight-induced
logit amplification.

Combined with the Frobenius-norm bound derived earlier, the bounded
logit range under Stiefel implies a positive lower bound on
\(\|J_{\mathrm{sm}}\|_F\). Concretely, if every logit lies in
\([-M, M]\) with \(M = R^2/\sqrt{d}\), then for every \(i\),
\[
p_i
=
\frac{e^{z_i}}{\sum_{\ell=1}^T e^{z_\ell}}
\ge
\frac{e^{-M}}{T e^M}
=
\frac{e^{-2M}}{T} .
\]

By the same argument applied to \(\max_i z_i\), every \(p_i\) is
also bounded above by \(1 - (T-1) e^{-2M}/T\), so each \(p_i\) lies
in \([a_T, 1-a_T]\) with \(a_T = e^{-2M}/T > 0\). Therefore
\[
\|J_{\mathrm{sm}}(z)\|_F^2
\ge
T\, a_T^2 (1-a_T)^2
=:
c(T,d,R)^2 ,
\]
which is independent of \(W_Q\) and \(W_K\). This is the lower bound
asserted in Proposition~\ref{prop:stiefel-nondeg}.

\paragraph{Scope of the argument.}
The preceding derivation should be interpreted as a mechanistic
explanation rather than a complete proof of training instability. The
mathematical statement is that DGram does not impose a uniform spectral
bound on the query and key projections. Equivalently, DGram admits
feasible sequences
\(\{W_Q^{(c)}, W_K^{(c)}\}_{c \to \infty}\) such that
\[
\operatorname{Off}\!\left((W_Q^{(c)})^\top W_Q^{(c)}\right)
=
\operatorname{Off}\!\left((W_K^{(c)})^\top W_K^{(c)}\right)
=0,
\qquad
\|W_Q^{(c)}\|_2 \|W_K^{(c)}\|_2 \to \infty .
\]
For suitable bounded inputs, the corresponding attention logits
\[
Z^{(c)}
=
\frac{X W_Q^{(c)} (W_K^{(c)})^\top X^\top}{\sqrt d}
\]
contain a row \(z^{(c)}\) whose logit gap satisfies
\[
\Delta_c
=
\max_i z_i^{(c)}
-
\max_{j \ne i^\star} z_j^{(c)}
\to \infty ,
\]
where \(i^\star = \arg\max_i z_i^{(c)}\). The softmax Jacobian bound
then gives
\[
\|J_{\mathrm{sm}}(z^{(c)})\|_F
\le
2(T-1)e^{-\Delta_c}
+
(T-1)^2 e^{-2\Delta_c}
\to 0 .
\]
Thus, DGram permits feasible attention weights along which the softmax
map becomes arbitrarily close to a one-hot routing map and its Jacobian
becomes arbitrarily small.

By contrast, Stiefel constraints remove this particular scale degree of
freedom. If
\[
W_Q^\top W_Q = I,
\qquad
W_K^\top W_K = I,
\]
then
\[
\|W_Q\|_2 = \|W_K\|_2 = 1,
\]
and hence the attention logits obey the uniform weight-induced bound
\[
\|Z\|_2
\le
\frac{\|X\|_2^2}{\sqrt d}.
\]
For token features satisfying \(\|x_i\|_2 \le R\), this also implies the
entrywise bound
\[
|Z_{ij}|
\le
\frac{R^2}{\sqrt d}.
\]
Therefore Stiefel attention rules out the specific mechanism in which
query and key weights grow in spectral norm and amplify attention logits
through the factor \(\|W_Q\|_2\|W_K\|_2\).

This argument does not imply that every DGram-attention training run
must become unstable, nor does it imply that Stiefel attention guarantees
non-degenerate gradients for all inputs or all optimization
trajectories. In a full Transformer, the attention logits also depend on
LayerNorm, residual connections, multiple heads, value and output
projections, optimizer state, and the data distribution. These effects
can modify the simplified single-head picture analyzed above.

The intended conclusion is therefore narrower: DGram allows a
weight-space direction along which attention logits can become
arbitrarily sharp, whereas Stiefel removes that weight-induced
amplification pathway. This contrast is consistent with our empirical
observation that configurations assigning DGram to attention exhibit
rapid singular-value growth and unstable training, while configurations
with Stiefel attention remain stable in our experiments.

\end{document}